# From Human Bias to Robot Choice: How Occupational Contexts and Racial Priming Shape Robot Selection

Jiangen He
jiangen@utk.edu
The University of Tennessee
Knoxville, Tennessee, USA

Wanqi Zhang
wzhang79@vols.utk.edu
The University of Tennessee
Knoxville, Tennessee, USA

Jessica Barfield
jessicabarfield@uky.edu
University of Kentucky
Lexington, Kentucky, USA

## Abstract

As artificial agents increasingly integrate into professional environments, fundamental questions have emerged about how societal biases influence human-robot selection decisions. We conducted two comprehensive experiments (N = 1,038) examining how occupational contexts and stereotype activation shape robotic agent choices across construction, healthcare, educational, and athletic domains. Participants made selections from artificial agents that varied systematically in skin tone and anthropomorphic characteristics. Our study revealed distinct context-dependent patterns. Healthcare and educational scenarios demonstrated strong favoritism toward lighter-skinned artificial agents, while construction and athletic contexts showed greater acceptance of darker-toned alternatives. Participant race was associated with systematic differences in selection patterns across professional domains. The second experiment demonstrated that exposure to human professionals from specific racial backgrounds systematically shifted later robotic agent preferences in stereotype-consistent directions. These findings show that occupational biases and color-based discrimination transfer directly from human-human to human-robot evaluation contexts. The results highlight mechanisms through which robotic deployment may unintentionally perpetuate existing social inequalities.

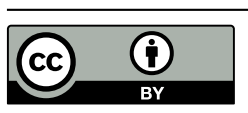

## 1 Introduction

The integration of robots into professional environments raises fundamental questions about how our human biases shape our interactions with artificial agents. As robots increasingly assume roles traditionally occupied by humans from healthcare assistants to construction workers, understanding the psychological mechanisms that influence robot selection and acceptance becomes critical. Within social science the stereotype content model has shown that professions and social groups are systematically evaluated along competence and warmth dimensions creating predictable associations between racial categories and occupational suitability [27]. Moreover, recent evidence suggests that humans project social categories upon robots, including racial and gender classifications that may mirror human biases such as discrimination, stereotypes, and racism in human-human interactions [1, 4, 5, 36]. Considering the racialization of technology, Benjamin [6] noted that there is a concern that its physical design may automate discrimination and bias. To investigate how biases may influence human-robot interaction (HRI) theories such as feminist theory, critical race theory, social role theory, and the stereotype content model provide useful conceptual frameworks to guide research and methodologies to explore biases in HRI [19, 37]. Extensive research in social science has demonstrated that stereotypes operate through automatic activation processes, where contextual cues prime specific cognitive schemas that influence subsequent judgments and decisions [3, 7]. For robots and other technologies implicit biases may manifest even when individuals consciously reject stereotypical thinking, as demonstrated by paradigms like stereotype threat, where priming racial identity can impair performance in stereotype relevant domains [8, 9, 11]. For research on robot biases occupational contexts can serve as particularly powerful stereotype activators, as social role theory suggests that stereotypes emerge from observable group membership across different occupational roles [10, 35].

From the literature it is also known that skin color is a strong cue which may trigger racism, discrimination, and stereotypes, with darker skin tones, consistently associated with lower status and reduced opportunities across multiple domains [2, 13, 16, 28, 30]. Given robots are often designed to appear as white or black, an emerging stream of research is beginning to show that robot skin color may trigger racism and biases during HRI [2, 22]. In the current research we used a more nuanced approach to manipulating robot color by varying the skin tone of the robot thus allowing us to produce a more human-appearing robot and to investigate whether our manipulation would trigger racism towards robots. Given the use of skin tone, critical race theory can be used to explore how systematic racism may be embedded in robot design and how it may influence the goal of achieving equity and fairness within society.

In addition, when robots are evaluated in terms of occupational contexts in which stereotype activation may shape robot choices for different occupations feminist theory may be useful to guide research on HRI given that unequal, undemocratic, or otherwise oppressive forces may be activated from the racialization of robot technology [38].

Despite this growing evidence base, key gaps remain in how occupational contexts and racial priming shape robot selection decisions. Most prior work emphasizes general bias measures, leaving interaction effects among participant race, task context, and robot human-likeness underexamined. Moreover, it remains unclear whether racial priming via human professional representations shifts subsequent robot selection, with direct implications for equitable deployment. Although social priming can influence perceptions of virtual agents [26], the transfer of occupational racial stereotypes to robot selection remains untested.

This study addresses these gaps through two experiments on robot selection across professional contexts. **Study 1** tests how task scenarios (construction, hospital, tutoring, sports) and participant characteristics shape selections among robots varying in skin tone and human-likeness. **Study 2** examines whether racial priming via human professional imagery shifts robot selection preferences. Together, these studies test context-dependent racial bias in robot selection and whether occupational stereotypes transfer from human–human to human–robot evaluation. We developed three research questions:

**RQ1:** How do professional task contexts influence robot skin tone selection patterns, and does robot human-likeness moderate these contextual effects?

**H1a:** Robot skin tone selections will differ across professional task contexts. **H1b:** The association between task context and robot skin tone selection will be stronger for more human-like robots than for less human-like robots.

**RQ2:** Does participant race moderate the relationship between task context and robot selection decisions?

**H2:** Participant race will moderate the relationship between task context and robot skin tone selection.

**RQ3:** How does racial priming through human professional representations systematically influence robot selection preferences, and does robot human-likeness amplify these priming effects?

**H3a:** Racial priming through human professional representations will shift robot skin tone selections relative to the non-primed condition. **H3b:** Priming effects on selection will be stronger for more human-like robots than for less human-like robots. **H3c:** Under stereotype-congruent primes, participants will be more likely to select robots whose skin tone is congruent with the racial cues activated by the prime.

## 2 Related Work

### 2.1 Priming and the Activation of Stereotypes

Priming refers to the incidental activation of knowledge structures, such as trait concepts and stereotypes, by the current situational context [3]. Decades of research demonstrate its powerful influence on cognition and behavior. For instance, Shih, Pittinsky, and Ambady showed that Asian American women's math performance improved when their Asian identity was primed but declined when their gender identity was made salient [31]. Steele and Aronson similarly found that priming African American identity before a test impaired academic performance, an effect known as stereotype threat [34]. Such findings illustrate that even subtle cues can activate stereotypes and bias performance and decision making. Within HRI, scholars have begun to recognize similar dynamics. Norouzi et al. found that social priming in virtual environments significantly influenced user attitudes toward virtual agents [26]. These studies suggest that priming is a critical mechanism through which biases may transfer to interactions with robots.

### 2.2 Occupational Contexts and Role-Based Stereotypes

Occupational roles are potent sources of stereotype activation. Social role theory argues that stereotypes emerge from observed occupational distributions [20], while the Stereotype Content Model shows systematic associations between social categories and occupational roles along competence and warmth dimensions [9, 27]. Empirical evidence reveals stereotypical linkages: Latinos with manual labor, Asians with academic competence, Blacks with athletic ability, and Whites with professional roles [9, 20, 28]. Task contexts can activate identity-linked stereotypes and influence performance [31], though little is known about how occupational scenarios prime stereotypes in HRI.

### 2.3 Social Biases in Human–Robot Interaction

Emerging literature demonstrates that humans project social categories onto robots. Participants display racial bias toward robots in shooter-bias tasks [1, 4], while robot skin color interacts with user prejudice to shape attributions of agency and experience [12]. Sparrow argues that race, as a social construction, extends to robots with significant ethical implications [32], while Howard and Borenstein emphasize how social inequities become embedded in AI systems, perpetuating bias [14]. Beyond race, nationality stereotypes also transfer to robots, influencing trust and suitability perceptions [10]. This literature indicates that HRI is not immune to social categorization mechanisms, making it relevant for examining priming effects.

### 2.4 Anthropomorphism and Human-Likeness

Anthropomorphism determines how strongly social categories influence user perceptions. While anthropomorphic design increases attributions of warmth, competence, and trust, it can also elicit discomfort near the human–nonhuman boundary [7, 21]. Physical human-likeness interacts with warmth and competence cues to shape service robot evaluations [5], while anthropomorphic features like skin tone interact with user prejudice to bias mind attribution judgments [12]. Robot appearance and accent influence nationality categorizations and trust evaluations [10], suggesting that anthropomorphism amplifies social category cue salience, making robots vulnerable to stereotype-based judgments when priming is present.

## 3 Methods

We conducted two complementary experiments examining how racial framing and professional contexts influence robot selection preferences. Study 1 investigated robot selection patterns under contextual exposure alone, presenting task scenarios without human figures to isolate environmental associations. Study 2 examined how racial priming through human professional imagery altered robot selection preferences. Participants from the United States were recruited through Prolific.com, with informed consent obtained prior to participation. Both studies employed identical robot stimuli that were systematically varied across skin tone and human-likeness dimensions. Participants received $1.80 compensation for completing either study. This research was approved by the institutional IRB.

## 3.1 Study 1: Robot Selection

*3.1.1 Experimental Design.* Study 1 employed a 4 (task scenario) × 5 (human-likeness level) mixed factorial design. Task scenario served as a within-subjects factor, while human-likeness level functioned as between-subjects factors. Participants were randomly assigned to one of five human-likeness (morphology) groups and completed all four task scenarios (see Figure 1).

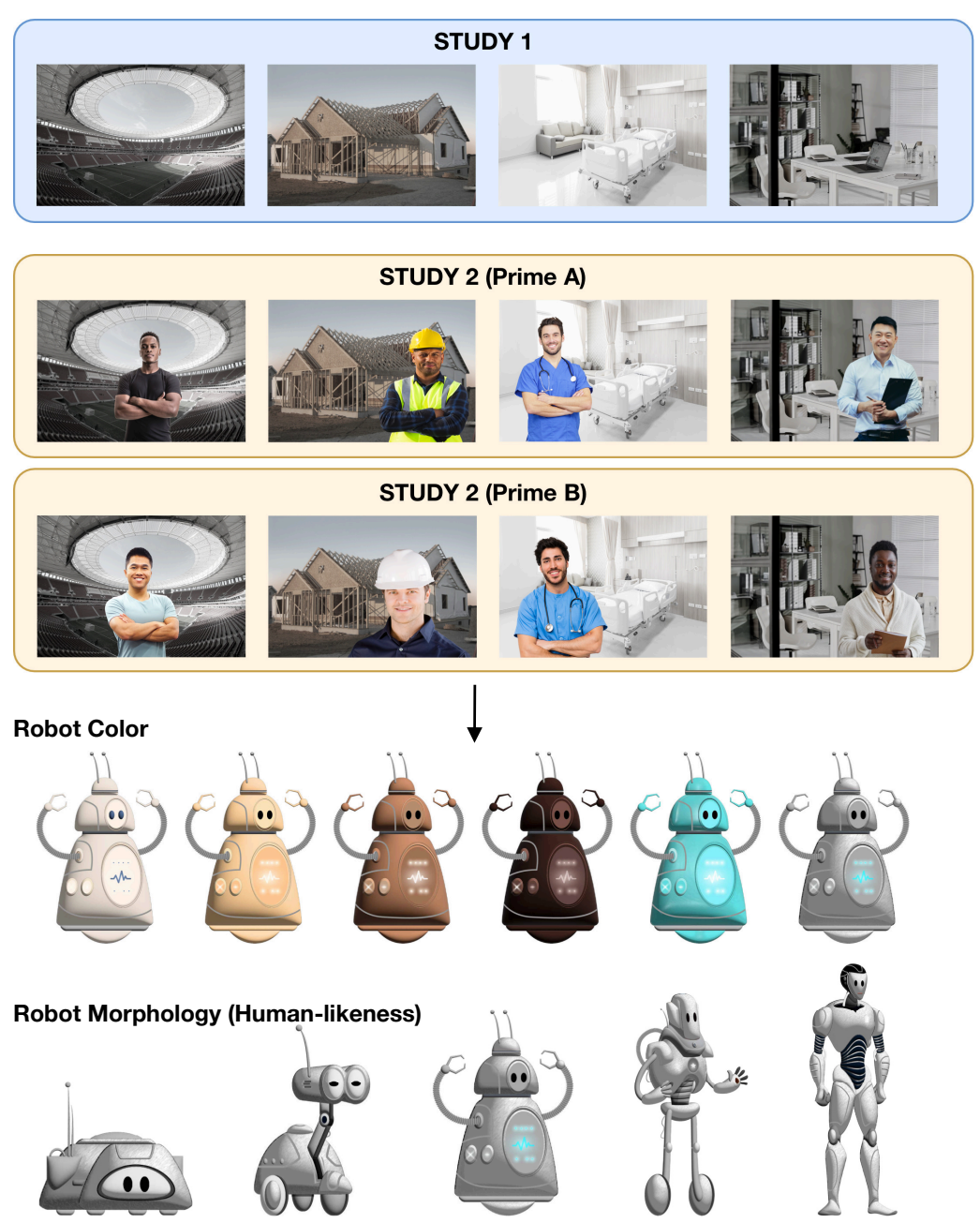

Figure 1: Experimental procedure, task scenarios, and robot stimuli. The robot stimuli vary by both colour and morphology, with morphology operationalized as five human-likeness levels. Appendix Figure 7

We treat task context as situational cues, and we manipulate them within subjects because they operate at the trial level and increase statistical power by controlling for stable individual differences. This design choice aligns with Social Role Theory and the Stereotype Content Model, which emphasize how contextual role cues structure expectations.

To reduce color-based confounds observed in pilot studies, participants viewed desaturated images of professional environments with each scenario. In each scenario, participants chose among six robots that varied in skin tone within their assigned human-likeness level. After each choice, participants rated helpfulness, deployment concerns, and the importance of human-like appearance. Post-experimental measures assessed general attitudes toward robots and AI and collected demographics (see Appendix B). Gender, age, and race were obtained via Prolific pre-screening.

*3.1.2 Task Scenarios.* Four professional contexts served as task scenarios: construction site, hospital, home tutoring, and sports field. These scenarios were strategically selected to represent real-world environments that vary systematically across critical dimensions including safety requirements, coordination complexity, time pressures, environmental conditions, and the balance between physical and cognitive work demands. Each task scenario provides a occupational context description, without mentioning race.

*3.1.3 Robot Stimuli.* Robot stimuli were systematically manipulated along two critical dimensions: human-likeness and skin color. We used rendered robot illustrations rather than photographs of real robots. Rendered stimuli allowed us to manipulate skin tone and span five Anthropomorphic Robot Database (ABOT)-based human-likeness levels while holding constant lighting, pose, scale, and surface materials across conditions [29]. Five representative designs were selected across a progression from mechanical to human-like: Level 1 featured immobile forms without limbs; Level 2 included robots with wheels and eyes; Level 3 incorporated manipulator arms; Level 4 added finger-like appendages and legged mobility; and Level 5 presented robots with full human contour and articulated limbs.

Skin tones were systematically generated using the Chicago Face Database [23]. Representative photographs from each racial group were sampled, and the dominant facial skin tone was extracted to determine the most representative color value for robot surfaces. Our research question isolates skin tone as the racial cue in order to establish whether this single dimension alone can trigger stereotype transfer. The final stimulus set included four distinct skin tones (light, medium, brown, dark) and two non-skin tones (silver, teal) as baseline conditions (see Appendix Figure 7). The non-skin tones provided a non-racialized comparison point and reduced the explicitness of forcing participants to choose among only skin-tone-like options. Except for these targeted manipulations, all other visual features of the robot stimuli remained constant to ensure that participant selections reflected only responses to human-likeness and color variations.

*3.1.4 Participants.* Following pilot studies ($N$ = 100), we recruited 421 participants for Study 1 (excluded participants with completion times under three minutes). We recruited Prolific participants using quota sampling, targeting a balanced distribution by gender (50/50) and race (25% per category). The final sample comprised 214 female (50.8%) and 207 male (49.2%) participants. The sample was evenly distributed across four racial categories: Black/African American (107, 25.4%), Latino/Hispanic (106, 25.2%), White (105, 24.9%), and Asian (103, 24.5%). Participant ages ranged from 18 to 83 years ($M$ = 35.7, $Md$ = 33.0, $SD$ = 13.0). With four task scenarios per participant, the study yielded 1,684 total observations. Participants completed the study averaging 758 seconds ($M$ = 12.6 mins, $Md$ = 10.8 mins).

## 3.2 Study 2: Robot Selection with Racial Priming

*3.2.1 Experimental Design.* Study 2 employed a 2 (priming condition) × 4 (task scenario) × 5 (human-likeness level) mixed factorial design. Both priming condition and task scenario served as within-subjects factors, while human-likeness level served as between-subjects factors. Robot stimuli and selection procedures mirrored those of Study 1, with participants choosing from six color options within their assigned human-likeness level.

The key experimental manipulation involved racial priming through human professional imagery integrated within task scenarios. The priming scenarios investigated how racial representations in professional contexts systematically shape robot selection decisions. Participants encountered four scenarios distributed across two distinct conditions: Prime A scenarios depicted professionals whose racial background corresponded with prevailing societal associations (stereotype-congruent racial representations), while Prime B scenarios showcased professionals from backgrounds less frequently linked to those occupational roles (stereotype-incongruent racial representations). Prime condition assignment was randomized with the constraint that each participant saw two scenarios under Prime A and two scenarios under Prime B. Scenario order was randomized.

The priming conditions were structured as follows:

- Construction Site (Prime A: Latino, Prime B: White)
- Hospital (Prime A: White, Prime B: Latino)
- Home Tutoring (Prime A: Asian, Prime B: Black)
- Sports Field (Prime A: Black, Prime B: Asian).

This counterbalanced design ensured that each racial group appeared in both stereotype-congruent and stereotype-incongruent conditions across different task scenarios, enabling comprehensive examination of stereotype activation effects.

*3.2.2 Participants.* We recruited 617 participants for Study 2. The same quota sampling method in Study 1 was used. The sample included 315 female (51.1%) and 302 male (48.9%) participants ranging from 18 to 81 years of age (M = 36.0, Md = 33.0, SD = 13.4). Participants were distributed evenly across four racial categories: Black/African American (162, 26.3%), White (160, 25.9%), Latino/Hispanic (154, 25.0%), and Asian (141, 22.9%). Task completion times reflected engagement levels comparable to Study 1 ($M$ = 12.6 mins, $Md$ = 10.4 mins). The study yielded 2,468 total observations across four task scenarios.

## 3.3 Measures and Analysis

*3.3.1 Dependent Variables.* Our primary dependent variable was an allocation decision, namely which robot a participant would assign to a given task scenario. This explicit, make-a-choice outcome mirrors consequential design and deployment decisions in downstream HRI workflows more directly than purely attitudinal or implicit measures. Three binary variables measured robot selection outcomes. *selected_{color}* coded selection of each specific skin-tone robot (0/1). *skin_tone_selection* coded selection of any skin-tone robot versus non-skin-tone alternatives (0/1). *priming_target* coded whether participants selected the robot color matching their racial priming condition (0/1).

*3.3.2 Independent Variables.* Demographic variables included participant *age* (continuous), *female* gender (0/1, male reference), ethnicity dummy variables for *black*, *latino*, and *white* (0/1, Asian reference).

Task perceptions included *task_helpfulness*, *task_concern*, and *task_importance* (all 7-point scales).

Post-experimental attitudes included perceived robot *human_likeness* (7-point scale), *color_ influence* on decisions (7-point scale), *robot_comfort* in daily life (11-point scale), and general *AI_opinion* (12-point scale).

Robot design variables included *task_scenario* with dummy variables for HOSPITAL, SPORTS, and TUTORING (0/1, CONSTRUCTION reference) and *human_likeness_group* assignment (1-5).

Priming variables included *priming_condition* (1=primed, 0=control), *priming_alignment* (whether selected robot matched prime target color), and *race_mirror* (whether participant race matched prime category).

*3.3.3 Statistical Analysis.* Data were analyzed using multilevel logistic regression models to account for the nested structure of participants completing multiple tasks. Random intercepts were included at the participant level to control for individual differences in selection preferences. Model fit was assessed using intraclass correlation coefficients (ICC), Akaike Information Criterion (AIC) , and pseudo-$R^2$ values. Significance testing was conducted using clustered standard errors to account for the non-independence of observations within participants.

# 4 Results

## 4.1 Study 1: Robot Selection Across Professional Contexts

*4.1.1 Robot Color Selection.* We first examine overall color selection patterns to identify color preferences in context-specific variations. Figure 2 demonstrates systematic variation in robot color selection across task contexts, with consistent preferences for non-skin tone options and distinct skin-tone patterns varying by professional context. Baseline robots (silver and teal) consistently achieved high selection rates across all scenarios, indicating a general preference for non-skin tones. These task-dependent differences in robot skin tone selection support **H1a**.

Task-specific skin-tone preferences revealed contextual biases [18]. Construction and sports scenarios demonstrated greater acceptance of darker skin tones, with dark robots receiving their highest selection rates in these contexts. Conversely, hospital and tutoring tasks showed strong preferences for lighter skin tones. Sports tasks presented relatively balanced skin-tone selection pattern. The hospital context exhibited the most polarized skin-tone preferences, with light robots achieving peak selection while dark robots received minimal selection.

*4.1.2 Human-likeness.* We investigated how robot human-likeness interacts with color preferences to understand whether human-like features amplify or diminish racial bias patterns. Figure 3 reveals interactions between robot human-likeness and color preferences across task contexts.

The most striking pattern emerges in the hospital scenario, where Light robot selection exhibits a dramatic peak at human-likeness level 3 (reaching approximately 42%), while Silver robot preferences show an inverse dip at the same level. This suggests that moderate human-likeness intensifies preferences for lighter, more human-like appearances in medical contexts. Tutoring and sports contexts show similarly dynamic patterns based on the quadratic trend lines (red curves), with skin-tone color selection peaking at moderate human-likeness levels (level 3). Construction context demonstrates more moderate variations overall, with baseline colors showing greater stability across human-likeness levels compared to skin-tone options. These patterns suggest that human-likeness may moderate racial and contextual biases in robot selection through complex, non-linear relationships. (Appendix C shows selection percentages for baseline vs skin-tone options across robot-humanlikeness levels.)

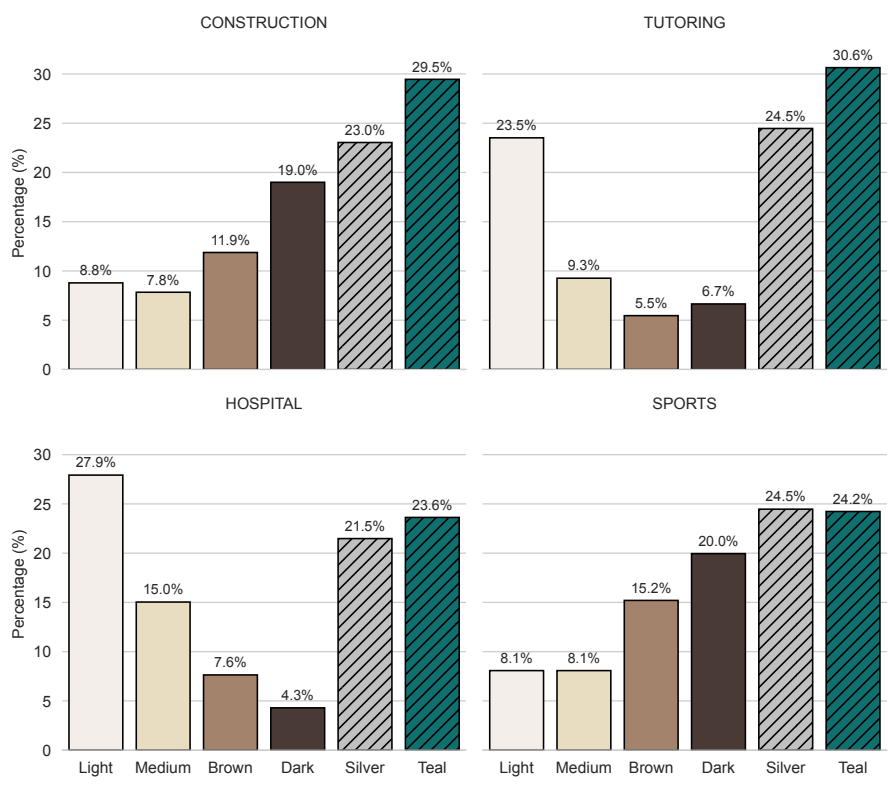

Figure 2: Robot color selection across ethnic groups and tasks. The two with hatch pattern are the two non-skin tones.

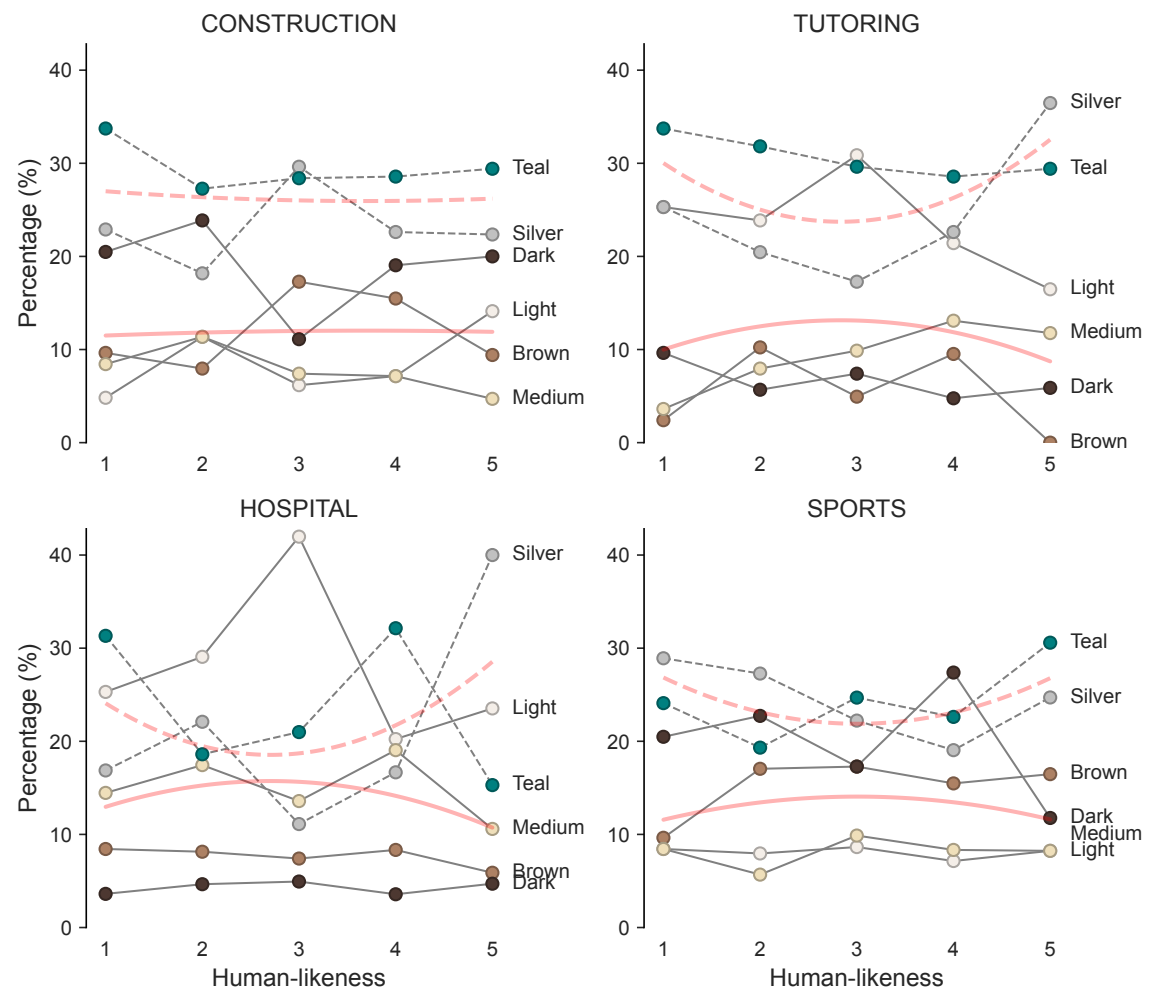

Figure 3: Robot color preferences by human-likeness level across tasks. Dashed lines show non-skin-tone color selection rates; solid lines show skin-tone color selection rates. Red curves display quadratic trend fits for each color category.

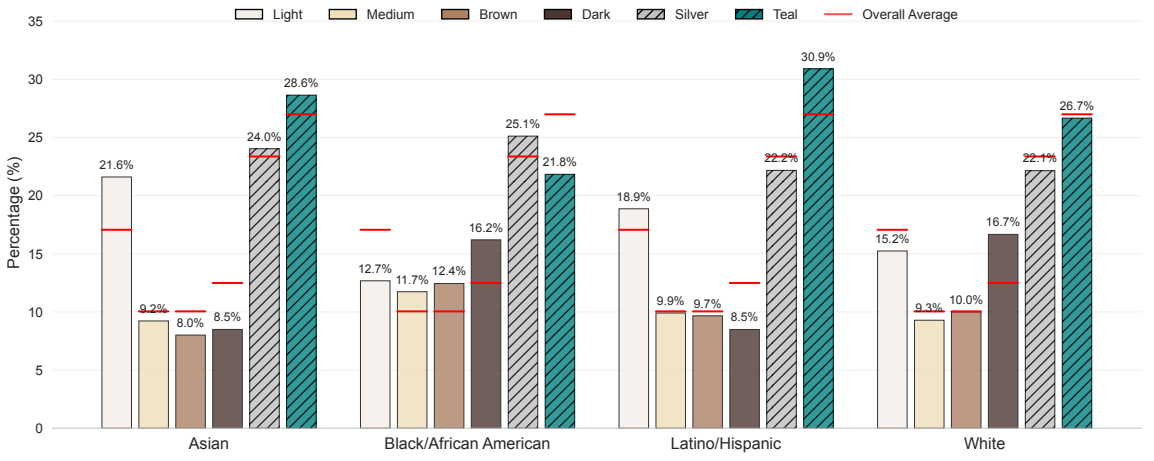

Figure 4: Robot color selection by participant races and tasks. The red lines are the overall average.

*4.1.3 Differences in Robot Selection by Participant Races.* We next examine whether participant race influences robot selection patterns to understand potential in-group preferences or bias effects. Figure 4 displays robot color selection by participant race, revealing distinct preferences across racial groups. Asian participants demonstrate a strong preference for Light robots (21.6%), significantly above the overall average, while also showing high selection rates for non-skin tone options (Silver and Teal). Black participants exhibit more balanced preferences across skin tone options, with notably higher selection rates for Dark robots compared to other groups. Latino participants show the strongest preference for Teal robots (30.9%). White participants display moderate preferences across most color categories with slightly elevated selection of Light robots. Notably, all groups consistently favor non-skin tone baseline options over human skin tone colors, suggesting a general preference for distinctly robotic appearances regardless of participant race.

Table 1 presents robot color selection frequencies by ethnicity and task, with cell colors indicating deviation from the mean (orange: above, blue: below; −10% to +10%). The table reveals several strong differences in specific colors within certain tasks.

Several distinct patterns emerge across participant races. Asian participants demonstrate consistent preference for light robots across care-oriented tasks, with selection rates of 30.1% in tutoring and 37.9% in hospital scenarios, both substantially above task means (23.5% and 27.9%). Black participants show the strongest preference for Dark robots in construction (31.8% vs. 19.0%) and sports (23.4% vs. 20.0% mean), suggesting comfort with darker robot appearances in physical labor contexts. White participants display notable variability, with significantly reduced Light robot selection in hospital scenarios (17.1% vs. 27.9% mean) while maintaining higher dark robot selection in construction tasks. These patterns suggest context-dependent decision-making involving implicit associations between skin tone and professional competency across different work domains. Overall, these race-contingent differences across task contexts are consistent with **H2**. However, these descriptive comparisons do not constitute a direct inferential test of moderation.

*4.1.4 Regression.* To quantify the relative contributions of demographic, attitudinal, and contextual factors on robot selection, we employ multilevel logistic regression models.

Table 2 presents five random intercept logistic regression models predicting overall skin-tone selection (**Skin**) and specific robot

Table 1: Robot color selection frequencies by race and task.

| Race | Light | Medium | Brown | Dark | Silver | Teal |
|---|---|---|---|---|---|---|
| CONSTRUCTION | | | | | | |
| Asian | 11.7% | 9.7% | 14.6% | 7.8% | 25.2% | 31.1% |
| Black | 4.7% | 8.4% | 12.1% | 31.8% | 23.4% | 19.6% |
| Latino | 7.5% | 7.5% | 10.4% | 9.4% | 24.5% | 40.6% |
| White | 11.4% | 5.7% | 10.5% | 26.7% | 19.0% | 26.7% |
| *Mean* | 8.8% | 7.8% | 11.9% | 19.0% | 23.0% | 29.5% |
| TUTORING | | | | | | |
| Asian | 30.1% | 8.7% | 1.0% | 3.9% | 28.2% | 28.2% |
| Black | 16.8% | 8.4% | 7.5% | 6.5% | 29.9% | 30.8% |
| Latino | 22.6% | 9.4% | 8.5% | 4.7% | 21.7% | 33.0% |
| White | 24.8% | 10.5% | 4.8% | 11.4% | 18.1% | 30.5% |
| *Mean* | 23.5% | 9.3% | 5.5% | 6.7% | 24.5% | 30.6% |
| HOSPITAL | | | | | | |
| Asian | 37.9% | 13.6% | 5.8% | 1.9% | 15.5% | 25.2% |
| Black | 21.0% | 19.0% | 8.6% | 2.9% | 28.6% | 20.0% |
| Latino | 35.8% | 13.2% | 8.5% | 3.8% | 17.9% | 20.8% |
| White | 17.1% | 14.3% | 7.6% | 8.6% | 23.8% | 28.6% |
| *Mean* | 27.9% | 15.0% | 7.6% | 4.3% | 21.5% | 23.6% |
| SPORTS | | | | | | |
| Asian | 6.8% | 4.9% | 10.7% | 20.4% | 27.2% | 30.1% |
| Black | 8.4% | 11.2% | 21.5% | 23.4% | 18.7% | 16.8% |
| Latino | 9.4% | 9.4% | 11.3% | 16.0% | 24.5% | 29.2% |
| White | 7.6% | 6.7% | 17.1% | 20.0% | 27.6% | 21.0% |
| *Mean* | 8.1% | 8.1% | 15.2% | 20.0% | 24.5% | 24.2% |

Cell colors show deviation from the mean (−10% (blue) to +10% (orange)).

Table 2: Random intercept logistic regression models for robot skin-tone selection.

| Predictor | Skin | Light | Med. | Brown | Dark |
|---|---|---|---|---|---|
| age | 1.00 | 1.00 | 1.00 | 1.01 | 0.98* |
| female | 0.83 | 0.82 | 1.31 | 1.17 | 0.83 |
| h_group | 0.94 | 1.00 | 1.04 | 1.04 | 0.91 |
| p_humanlike | 0.93 | 1.01 | 0.93 | 1.03 | 1.02 |
| color_influence | 0.88 | 1.03 | 0.84 | 0.93 | 1.25 |
| robot_comfort | 1.06 | 0.98 | 0.95 | 1.28 | 0.91 |
| ai_opinion | 1.01 | 1.19 | 1.15 | 0.84 | 0.79 |
| helpful | 1.13 | 1.29 | 0.73* | 0.74* | 1.36* |
| concern | 1.01 | 1.02 | 0.87 | 1.13 | 1.12 |
| importance | 1.04 | 0.98 | 0.96 | 1.03 | 1.23 |
| task_hospital | 1.33 | 4.43*** | 2.14** | 0.51* | 0.12*** |
| task_sports | 1.14 | 0.83 | 1.01 | 1.32 | 0.97 |
| task_tutoring | 0.88 | 4.99*** | 1.39 | 0.43** | 0.26*** |
| race_black | 1.23 | 0.38*** | 1.43 | 1.49 | 1.57 |
| race_latino | 0.97 | 0.73 | 1.10 | 1.41 | 0.98 |
| race_white | 1.15 | 0.47** | 1.05 | 1.12 | 2.19** |
| ICC | 0.26 | 0.54 | 0.54 | 0.49 | 0.46 |
| AIC | 2339.7 | 967.8 | 849.5 | 831.4 | 840.6 |
| Pseudo $R^2$ | 0.01 | 0.13 | 0.03 | 0.05 | 0.14 |
| N | 1682 | 835 | 835 | 835 | 835 |

Note: *p < 0.05, **p < 0.01, ***p < 0.001.
Odds ratios shown with significance levels.
DVs are skin-tone (Skin), light (Light), medium (Med.), brown (Brown), and dark (Dark).

color choices (**Light**, Medium (**Med.**), **Brown**, **Dark**). The models demonstrate substantial between-participant heterogeneity in robot selection preferences, with intraclass correlation coefficients ranging from 0.26 (overall skin-tone selection) to 0.54 (Light and Medium robot models). This indicates that 26-54% of total variance occurs between rather than within participants, validating the multilevel modeling approach.

Model explanatory power varies substantially across robot color categories. The Dark robot model achieves the highest pseudo $R^2$ (0.14) with the lowest AIC (840.6), followed by Light robots (pseudo $R^2$ = 0.13, AIC = 967.8). Medium and Brown robot models demonstrate weaker fit (pseudo $R^2$ = 0.03 and 0.05 respectively), suggesting that intermediate skin-tone preferences are less predictable. The overall skin-tone selection model shows negligible explanatory power (pseudo $R^2$ = 0.01, AIC = 2339.7), indicating limited ability to explain variation in selecting skin-tone versus baseline colors.

Gender and robot human-likeness (*h_group*) exhibit no significant effects across any model. Accordingly, we do not find support for **H1b** in the regression results, even we observe the moderate effects of humanlikeness in Figure 3. Age demonstrates minimal but significant effects, with each additional year reducing Dark robot selection odds by 2%. Among attitudinal variables, only task-specific helpfulness demonstrates significant effects on selecting medium, brown, and dark robots.

Hospital and tutoring contexts produce the largest and most consistent effects across all models. Hospital tasks increase Light robot selection odds by 343% (OR = 4.43, $p < 0.001$) and Medium robot odds by 114% (OR = 2.14, $p < 0.01$), while reducing Dark robot odds by 88% (OR = 0.12, $p < 0.001$). Tutoring tasks increase Light robot odds by 399% (OR = 4.99, $p < 0.001$), while reducing Brown robot odds by 57% (OR = 0.43, $p < 0.01$) and Dark robot odds by 74% (OR = 0.26, $p < 0.001$). Sports tasks show no significant effects across any model, with Construction serving as the reference category. Task context variables produced the largest effect sizes, with Hospital (OR = 4.43) and Tutoring (OR = 4.99) scenarios showing 4.4-fold and 5.0-fold increases in Light robot selection odds, respectively. These statistical evidences support **H1a**.

Participant race demonstrates significant effects primarily for Light and Dark robot selection. Black participants show 62% reduced odds of Light robot selection (OR = 0.38, $p < 0.001$) compared to Asian participants. White participants demonstrate 53% reduced odds of Light robot selection (OR = 0.47, $p < 0.01$) and 119% increased odds of Dark robot selection (OR = 2.19, $p < 0.01$). Latino participants show no significant effects across any robot color category compared to Asian.

## 4.2 Study 2: Racial Priming Effects on Robot Selection

*4.2.1 Stereotype-Congruent vs. Stereotype-Incongruent Priming.* We examine whether exposure to human professionals of specific races influences subsequent robot selection patterns through stereotype activation mechanisms. Figure 5 demonstrates the differential impact of racial priming on robot color selection across four professional task scenarios. The comparison between Prime A (stereotype-congruent) and Prime B (stereotype-incongruent) conditions reveals systematic shifts in selection preferences that vary by task context.

In construction scenarios, Prime A conditions (featuring Latino professionals) increased selection of darker-toned robots compared to Prime B conditions (featuring White professionals). However, Prime B conditions did not increase light robot selection compared to Prime A conditions. Similarly, in hospital scenarios, Prime A conditions (featuring White professionals) increased selection of lighter-toned robots relative to Prime B conditions (featuring Latino professionals). The tutoring context exhibited pronounced priming effects, with Prime A conditions (featuring Asian professionals) substantially increasing selection of medium-toned robots compared to Prime B conditions (featuring Black professionals). Sports scenarios demonstrated strong priming effects, with Prime A conditions (featuring Black professionals) showing increases in dark robot selection compared to Prime B conditions (featuring Asian professionals).

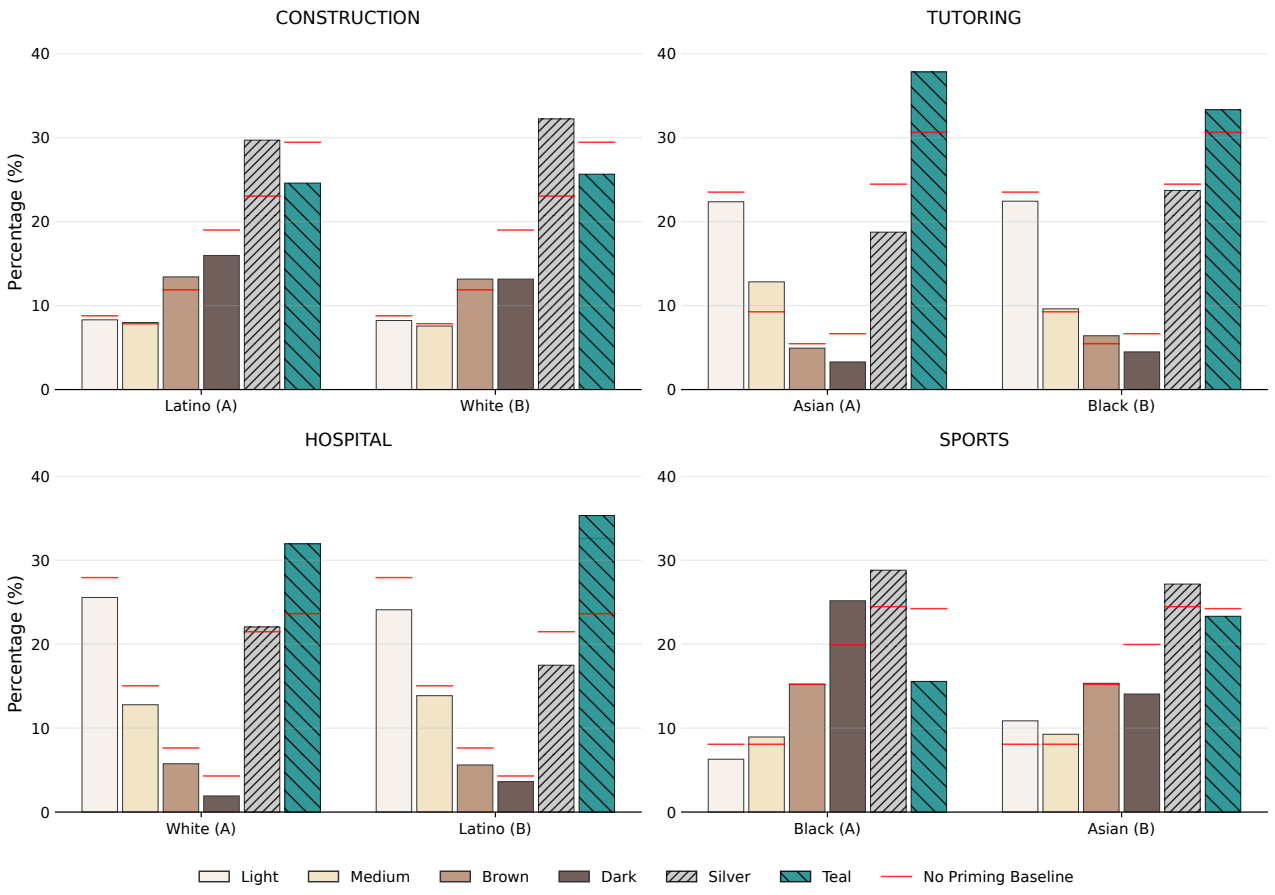

Figure 5: Robot color selection patterns under racial priming across tasks. Compares Prime A (stereotype-congruent) versus Prime B (stereotype-incongruent) conditions across construction, hospital, tutoring, and sports contexts. Red lines show baseline selection rates without priming.

These patterns indicate that racial priming operates through context-specific stereotype activation, where the presence of professionals from different racial backgrounds influences subsequent robot selection decisions in ways that align with broader societal associations between race and occupational roles. Overall, these priming-induced shifts in robot skin tone selection support **H3a**.

*4.2.2 Prime-Matched Robot Selection.* We examined whether priming effects vary by participant race and analyze selections of robots matching the racial characteristics of priming stimuli. Figure 6 displays prime-matched robot selection by participant race and task scenario. This analysis focuses on selections of skin-tone robots matching the ethnic priming condition (White → Light, Asian → Medium, Latino → Brown, Black → Dark). For example, in construction scenarios with Prime A (Latino) and Prime B (White) conditions, we compared Brown robot selection between no priming and Prime A conditions and Light robot selection between no priming and Prime B conditions. Selection rates are calculated as percentages of all skin-tone robot selections to account for changes in non-skin-tone robot selection under different priming conditions.

Overall, Prime A conditions show larger effects than Prime B conditions. Two Prime B conditions demonstrate negative effects on skin-tone robot selection: Black professionals in tutoring and Latino professionals in hospital contexts. In other task scenarios, Prime A condition effects exceed Prime B condition effects. Beyond overall trends, priming effects vary by participant race. For example, the negative effect of Prime B conditions (featuring Black professionals) in tutoring tasks did not affect Black participants, while Asian participants were more likely affected by Prime B conditions (featuring Asian professionals) in sports tasks compared to other races.

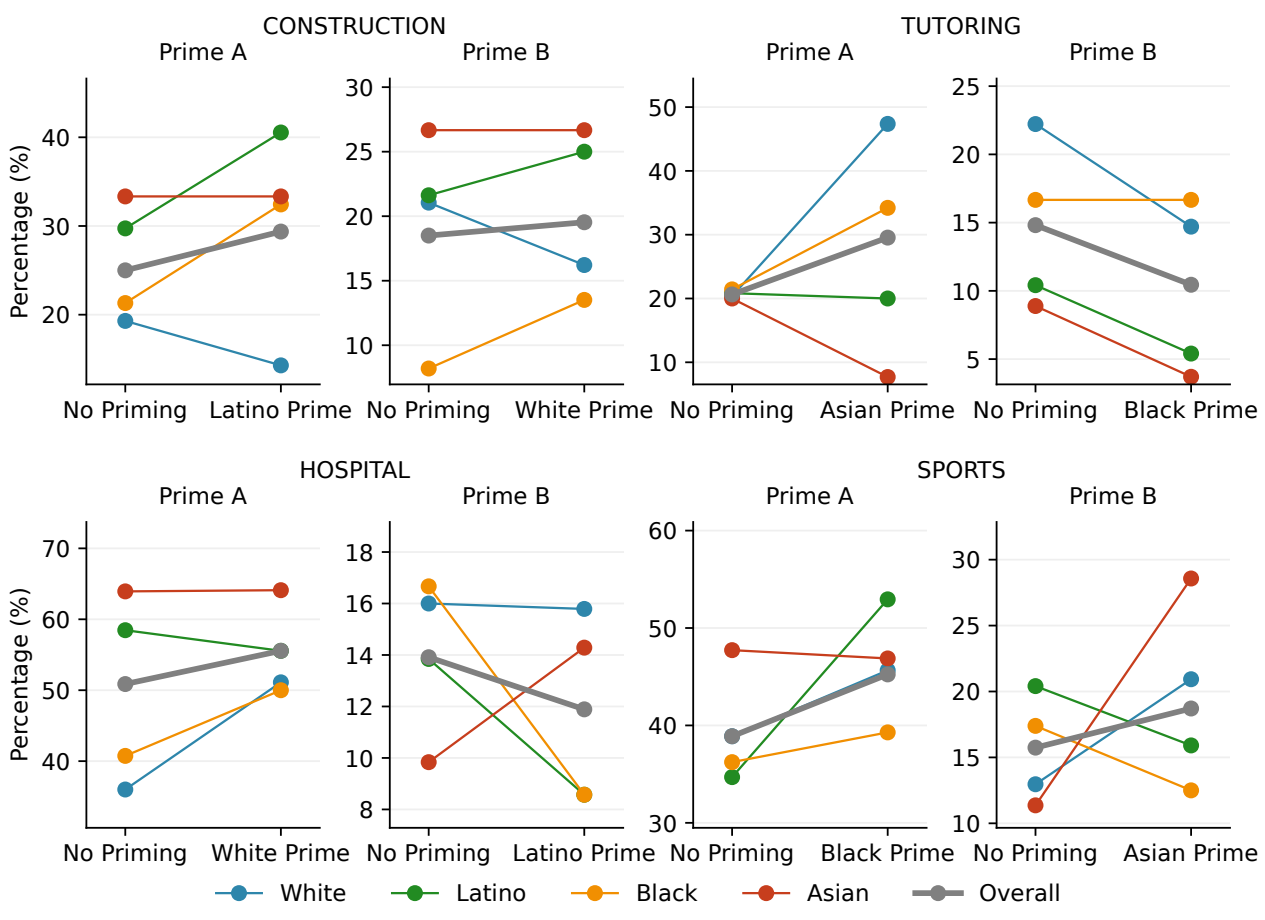

Figure 6: Prime-matched robot selection by participant race and task. Shows percentage of participants selecting robots matching the priming condition. Lines connect baseline (left) to priming conditions (right) for each ethnicity and overall average (grey). Y-axis shows prime-matched selections as percentage of all skin-tone selections. Note: Y-ranges vary by subplot.

*4.2.3 Regression.* Table 3 presents a multilevel logistic regression model predicting prime-matched target selection. In this model, we investigate whether stereotype priming influences participants more than stereotype-incongruent priming. The dependent variable is *priming_target*, which indicates whether participants selected the robot color matching their racial priming condition. For example, if the prime is a latino professional (including both Prime A and B), the dependent variable is 1 if the participant selected the Brown robot, and 0 otherwise.

The results reveal a strong priming effect, with stereotype priming (Prime A) increasing the odds of selecting a stereotype-congruent robot by a factor of 5.93 (OR = 5.93, $p < 0.001$), corresponding to a 493% increase relative to baseline odds. This stereotype-congruent increase in prime-matched selections supports **H3c**. This represents the largest effect size in the model, with participants 5.93 times more likely to select robots matching the racial characteristics of human professionals featured in their assigned task scenarios. Among task contexts, hospital scenarios significantly increased the likelihood of prime-matched selections compared to the construction reference category. The model explained 9% of the variance in the selections (pseudo $R^2$ = 0.09) with high intraclass correlation (ICC = 0.57), indicating that 57% of the total variance occurred between participants rather than within participants. Neither participant race, gender, robot human-likeness, nor attitudes toward robots significantly predicted the selections. The absence of significant interaction effects between stereotype priming and human-likeness variables indicates that priming influences remained consistent across different levels of robot human-likeness. Accordingly, **H3b** was not supported.

Table 3: Random intercept logistic regression predicting prime-matched robot selection.

| Predictor | OR | Predictor | OR | Predictor | OR |
|---|---|---|---|---|---|
| ***Demographics*** | | ***Task (vs. Construction)*** | | ***Attitudes*** | |
| age | 1.01 | Hospital | 1.52* | humanlikeness | 1.12 |
| female | 1.03 | Sports | 1.46 | color influence | 0.94 |
| h_group | 0.93 | Tutoring | 0.78 | robot comfort | 1.06 |
| ***Priming*** | | ***Race (vs. Asian)*** | | AI opinion | 0.99 |
| stereotype_ priming | 5.93*** | Black | 0.81 | helpful | 0.97 |
| race_match | 1.02 | Latino | 0.91 | concern | 0.93 |
| | | White | 0.86 | importance | 1.08 |
| ***Interactions*** | | | | | |
| stereotype × humanlikeness | | 0.89 | | stereotype × h_group | 1.01 |
| *Model Fit: ICC = 0.57, Pseudo $R^2$ = 0.09* | | | | | |

*p<0.05, **p<0.01, ***p<0.001. Odds ratios shown with significance levels. DV is ***priming_target***.

## 5 Discussion and Conclusions

This study provides systematic evidence that occupational contexts and racial priming significantly influence robot selection decisions, mirroring biases observed in human-human interactions. Through two experiments with 1,038 participants, we demonstrate that human-robot interaction is not immune to discriminatory mechanisms, with profound implications for equitable robot deployment. Notably, the most frequently selected colours were the non-human baseline options (silver and light-teal), rather than any skin-tone.

*Professional Task Contexts and Robot Selection Patterns (RQ1).* Hospital and tutoring scenarios exhibited pronounced favoritism toward lighter-skinned robots, with 343-399% increased odds compared to construction tasks. These preferences reflect societal associations between lighter skin tones and perceived competence in care-oriented professions [9]. Conversely, construction and sports scenarios showed greater tolerance for darker-skinned robots, mirroring occupational segregation patterns [17, 28]. These findings align with social role theory predictions that stereotypes emerge from observed occupational distributions [20].

Robot human-likeness revealed non-linear moderation effects peaking at moderate anthropomorphism levels. Light robot selection reached 42% at human-likeness level 3 in hospital scenarios, suggesting that moderate human-likeness intensifies skin-tone preferences by making racial categorization salient without triggering uncanny valley effects [7, 25]. This pattern suggests a critical threshold where robots become sufficiently human-like to activate social categorization processes while remaining comfortable for users. The smaller the percentage differentiation between skin-tone color when human-likeness is high indicates the participants are more cautious about the racial categorization.

The human-likeness findings have important theoretical and practical implications. Theoretically, they support the view that racial bias in HRI emerges from the same social cognitive mechanisms that govern human-human interactions—namely, the automatic categorization of entities that appear sufficiently human-like [5]. The inverted-U relationship between anthropomorphism and bias suggests that as robots become more human-like, they increasingly trigger stereotype-based evaluations until reaching very high human-likeness levels where other factors may intervene. Practically, these findings present a design dilemma: while moderate human-likeness may enhance user engagement and trust [7], it also amplifies susceptibility to discriminatory preferences.

*Participant Race and Moderation Effects (RQ2).* Black participants showed 62% reduced odds of selecting light robots and 133% increased preference for dark robots compared to Asian participants, suggesting resistance to societal preferences for lighter skin tones and positive identification with darker-skinned artificial agents [15, 24]. Asian participants exhibited strong preferences for light robots, particularly in care-oriented tasks, potentially reflecting cultural skin color hierarchies [8]. White participants displayed context-sensitive preferences aligning with task-appropriate stereotypes, while Latino participants showed the most neutral patterns, preferring non-skin-tone options.

*Racial Priming and Stereotype (RQ3).* Stereotype-congruent priming produced strong effects, increasing odds of selecting matching robot colors by 493% (OR = 5.93, $p < 0.001$). This represents one of the largest effect sizes observed, demonstrating powerful influence of contextual racial cues on technological preferences [3]. Priming reduced overall skin-tone robot selection by 91-95% while concentrating selections among stereotype-consistent options, indicating both heightened awareness of racial categories and preference for avoiding racialized choices.

*Broader Implications.* The systematic transfer of occupational stereotypes to robot selection risks creating technological segregation where lighter-skinned robots dominate prestigious professions while darker-skinned robots are relegated to manual labor roles. This could reinforce harmful stereotypes and influence career aspirations [34]. The finding that moderate human-likeness intensifies bias suggests complex design trade-offs between anthropomorphic appeal and discrimination amplification [5]. The 493% increase in stereotype-congruent selections following racial priming demonstrates that environmental context significantly shapes technological preferences. The systematic bias raises questions about algorithmic fairness requiring expanded anti-discrimination frameworks for technological representations [14, 33]. From a design perspective, reducing or avoiding skin-tone cues may lower the salience of racial categorization, but it may not fully resolve the issue if users infer race from other appearance signals or contextual framing. Future work should test whether "race-free" color palettes prevent racialized inferences, and identify which non-color cues most strongly drive perceived race in HRI. Our finding that participants consistently preferred non-skin-tone options suggests that offering neutral colours can mitigate the risk of perpetuating racial bias in robot selection.